\documentclass[journal]{IEEEtran}
\ifCLASSINFOpdf
\else
\fi

\usepackage{url} 
\usepackage{graphicx} 
\usepackage[utf8]{inputenc} 
\usepackage[T1]{fontenc}    
\usepackage{booktabs}       
\usepackage{amsfonts}       
\usepackage{nicefrac}       

\usepackage{subfiles}
\usepackage{amssymb}

\usepackage{algorithm}
\usepackage{algorithmic}
\usepackage[algo2e,ruled,linesnumbered]{algorithm2e}
\usepackage{multirow}
\usepackage{bm}
\usepackage{amsmath}
\usepackage{amsthm}
\usepackage{makecell}

\usepackage{comment}
\usepackage{color}
\usepackage{etoolbox}
\newbool{inccomment}
\setbool{inccomment}{true}

\usepackage[english]{babel}
\newtheorem{theorem}{Theorem}

\usepackage{caption}

\newcommand{\hl}[1]{\ifbool{inccomment}{{\color{magenta}#1}}{}}
\newcommand{\ww}[1]{\ifbool{inccomment}{{\color{blue} #1}}{}}
\newcommand{\yc}[1]{\ifbool{inccomment}{{\color{red} #1}}{}}
\newcommand{\sout}{\textit{SmoothOut}}
\newcommand{\ssout}{\textit{Stochastic SmoothOut}}
\newcommand{\alexnet}{\textit{AlexNet}}

\newcommand{\ie}{\textit{i.e.}}
\newcommand{\etal}{\textit{et al.}}

\newcommand{\df}{\mathcal{D}(\bm{w}_f, \tau)}
\newcommand{\ds}{\mathcal{D}(\bm{w}_s, \varepsilon)}

\begin{document}
%
\title{\sout{}: Smoothing Out Sharp Minima \\ to Improve Generalization in Deep Learning}
%
%
%

\author{Wei~Wen,
        Yandan~Wang,
        Feng~Yan,~\IEEEmembership{Member,~IEEE},
        Cong~Xu, 
        Chunpeng~Wu, \\
        Yiran~Chen,~\IEEEmembership{Fellow,~IEEE}
        and~Hai~(Helen)~Li,~\IEEEmembership{Fellow,~IEEE}
\thanks{W. Wen is with the Department
of Electrical and Computer Engineering, Duke Univerisity, Durham,
NC, 27708 USA e-mail: (see www.pittnuts.com).}
\thanks{Y. Wang is with University of Pittsburgh.}
\thanks{F. Yan is with University of Nevada -- Reno.}
\thanks{C. Xu is with HP Labs.}
\thanks{C. Wu, H. Li and Y. Chen are with Duke Univerisity.}
\thanks{Manuscript for review.}}

%
%

\markboth{Preprint}%
{Wen \MakeLowercase{\textit{et al.}}: SmoothOut: Smoothing Out Sharp Minima to Improve Generalization in Deep Learning}
%



\maketitle

\begin{abstract}
In Deep Learning, \textit{Stochastic Gradient Descent} (SGD) is usually selected as a training method because of its efficiency; however, recently, a problem in SGD gains research interest: sharp minima in Deep Neural Networks (DNNs) have poor generalization; especially, large-batch SGD tends to converge to sharp minima. It becomes an open question whether escaping sharp minima can improve the generalization. 
To answer this question, we propose \sout{} framework to smooth out sharp minima in DNNs and thereby improve generalization.
In a nutshell, \sout{} perturbs multiple copies of the DNN by noise injection and averages these copies.
Injecting noises to SGD is widely used in the literature, but \sout{} differs in lots of ways: 
(1) a de-noising process is applied before parameter updating; 
(2) noise strength is adapted to filter norm;
(3) an alternative interpretation on the advantage of noise injection, from the perspective of sharpness and generalization;
(4) usage of uniform noise instead of Gaussian noise.
We prove that \sout{}  can eliminate sharp minima.
Training multiple DNN copies is inefficient, we further propose an unbiased stochastic \sout{} which only introduces the overhead of noise injecting and de-noising per batch. An adaptive variant of \sout{}, \textit{Ada}\sout{}, is also proposed to improve generalization.
In a variety of experiments, \sout{} and \textit{Ada}\sout{} consistently improve generalization in both small-batch and large-batch training on the top of state-of-the-art solutions. 
\end{abstract}


%
\IEEEpeerreviewmaketitle


%

\section{Introduction}

\label{sec:intro}
\textit{Stochastic Gradient Descent}~(SGD) is the dominant optimization method used to train \textit{Deep Neural Networks}~(DNNs). 
However, the generalization of DNNs needs more understanding. 
Recently, one observation is that large-batch SGD has worse generalization than small-batch SGD \cite{diamos2016persistent}\cite{keskar2016large}\cite{goyal2017accurate}.
The accuracy difference between small-batch training and large-batch training is the well known ``generalization gap'' \cite{keskar2016large}. 

Reasons behind the ``generalization gap'' are still under active research. 
Hoffer~\etal{}~\cite{hoffer2017train} hypothesizes that the process of SGD is similar to ``random walk on a random
potential'' \cite{bouchaud1990anomalous}. This hypothesis attributes generalization gap to the limited number of parameter updates, and suggests to train more iterations. Learning Rate Scaling (LRS) was also proposed to match walk statistics  to close the gap.
Inspired by this hypothesis, practical techniques are proposed \cite{goyal2017accurate}\cite{smith2017don}\cite{you2017scaling}\cite{akiba2017extremely}. 

Another appealing hypothesis, which arouses recent research interest, is that the generalization is attributed to the flatness of minima \cite{hochreiter1997flat}\cite{chaudhari2016entropy}; that is, flat minima have good generalization while sharp minima can worsen it.
The hypothesis can be applied to both small-batch and large-batch SGD, but large-batch SGD tends to converge to sharper minima, ending up with the generalization gap.
Sharp minima have bad generalization due to their high over-fitting to training data \cite{hochreiter1997flat}\cite{rissanen1978modeling} and high sensitivity to noises \cite{chaudhari2016entropy}.

Jastrz{\k{e}}bski~\etal{}~\cite{jastrzkebski2018finding} showed the connection between these two hypotheses: LRS motivated by ``random walk'' leads to flatter minima and helps to improve the generalization.
Our approach is based on the second hypothesis, targeting on escaping sharp minima for better generalization in both small-batch and large-batch SGD.
Moreover, our approach can enhance techniques inspired by the first hypothesis and \textit{further} improve generalization.

Keskar~\etal{}~\cite{keskar2016large} attempted to escape sharp minima through data augmentation, conservative training and adversarial training. However, all trials ``do not completely remedy the problem'' \cite{keskar2016large}, leaving how to avoid sharp minima as an open question.
We propose \textit{SmoothOut} to smooth out sharp minima and guide the convergence of SGD to flatter regions.
\textit{SmoothOut} slightly perturbs DNN \textit{function} by noise injecting or function reshaping, then averages all perturbed DNNs. 
Because sharp minima are sensitive to perturbation, slight perturbation can result in significant function increase at each sharp minimum, which means the averaged value will be high. In this way, sharp minima can be eliminated. 
Conversely, small perturbation only influences the margin of each flat region and the ``flat bottom'' still aligns well with the original ``bottom''. Averaging aligned ``bottoms'' can maintain the original minimum.
Beyond this intuition, we prove that \sout{} under uniform noises can eliminate sharp minima while maintaining flat minima.
Note that we majorly use uniform noise for study as it is well motivated, but other noise types like Gaussian noise can fit into our new \sout{} framework.
Moreover, training over many perturbed DNNs for averaging is computation intensive. We propose \textit{Stochastic SmoothOut}, which injects noise per iteration during SGD. We prove that \textit{Stochastic SmoothOut} is equivalent to the original \textit{SmoothOut} in expectation. 
Adaptive \sout{} -- \textit{Ada}\sout{}, is also proposed to further improve generalization by adapting noise strength to filter norm.
Our experiments show that \textit{SmoothOut} and \textit{Ada}\sout{} can help to escape sharp minima and improve the generalization. \textit{SmoothOut} and \textit{Ada}\sout{} are easy to be implemented and our code is at \url{https://github.com/wenwei202/smoothout}.

\section{Related work}
\label{sec:related}
\textbf{Sharp Minima and Generalization.}
Why deep neural networks generalize well still needs deeper understanding \cite{zhang2016understanding}. 
As aforementioned, one hypothesis is that SGD finds flat minima which can generalize well \cite{hochreiter1997flat}\cite{keskar2016large}.
As Hochreiter~\etal{}~\cite{hochreiter1997flat} pointed out, based on Minimum Description (Message) Length \cite{rissanen1978modeling}\cite{wallace1968information} theory, a flatter minimum can be encoded in fewer bits which indicates a simpler DNN model for better generalization. Alternative explanations are based on Bayesian learning \cite{mackay1992practical}\cite{smith2017bayesian}. 
Dinh~\etal{}~\cite{dinh2017sharp} further argue that current definitions of sharpness are problematic and redefinition is required for explanation. 
However, Keskar~\etal{}~\cite{keskar2016large} indeed found that large-batch training sticks to sharp minima and has bad generalization.
Different from previous work, we focus on new variants of SGD to escape sharp minima. We find that our method not only can escape sharp minima during large-batch training but also guide small-batch training to flatter ones, therefore, improving generalization in both cases.
Chaudhari~\etal{}~\cite{chaudhari2016entropy} proposed Entropy-SGD which maximizes local entropy to bias SGD to flat minima (``wide valleys''). Local entropy was constructed by building connections between Gibbs distribution and optimization problems.
The gradients of local entropy was estimated by Langevin dynamics \cite{welling2011bayesian}, which is computation intensive. 
Compared with Entropy-SGD, our \textit{SmoothOut} is more efficient since noise injection and de-noising is the only overhead. This enables \textit{SmoothOut} to scale to larger dataset like ImageNet \cite{deng2009imagenet}. Moreover, \textit{SmoothOut} consistently improves generalization in all experiments, while Entropy-SGD achieved ``comparable generalization error''. 
Izmailov~\etal{}~\cite{izmailov2018averaging} proposed \textit{Stochastic Weight Averaging}~(SWA), which records the newest parameter points along the trajectory of SGD and then simply averages them to get the final optimum. 
Comparing with SWA, \sout{} performs stochastic averaging over perturbed models; SWA relies on a pre-training to converge near to minima, while \sout{} can train from scratch; moreover, it is unanswered if SWA can improve generalization in large-batch SGD, while \sout{} can improve it both experimentally and theoretically.

\textbf{Noise Injection.}
Noise injection is a commonly used method in SGD \cite{neelakantan2015adding}\cite{mobahi2016training}\cite{fortunato2017noisy}\cite{plappert2017parameter}\cite{li2016whiteout}\cite{ho2008weight} and Bayesian neural networks \cite{neal2012bayesian}\cite{hinton93keeping}\cite{kingma2013auto}, where usually Gaussian noises are injected to parameters or gradients for exploration or distribution approximation. 
Differently, our method is motivated by eliminating sharp minima, which leads to some key differences:  
(1) a de-noising process is applied before parameter updating; 
(2) noise strength is adapted to filter norm;
(3) an alternative interpretation on the advantage of noise injection, from the perspective of sharpness and generalization;
(4) usage of uniform noise instead of Gaussian noise.
Our experiments will show that uniform noise is superior to Gaussian noise.
Moreover, our \sout{} framework is agnostic noise types
Any noise type can fit to the framework and may achieve the goal. 
We adopt uniform noise as a major study because it is well motivated as will be shown in Section~\ref{sec:method}.
Dropout \cite{srivastava2014dropout} is a popular method to avoid over-fitting and include uncertainty \cite{gal2016dropout} by randomly drop neurons, however, large-batch training with Dropout still has the generalization gap as shown experimentally.
The reason is: as Keskar~\etal{}~\cite{keskar2016large} observed, sharp region only expands in a small dimensional subspace and most directions are flat; however, Dropout only perturbs a subspace such that the sharp directions cannot be frequently perturbed; conversely, our method effectively perturbs the whole space including sharp direction. We will explain the connections between Dropout and our method.

\textbf{Large-batch SGD.}
Large-batch SGD is a loosely related work because \sout{} is a general SGD approach.
However, as sharp minima in large-batch SGD become severer \cite{keskar2016large} and accuracy loss is generally observed, an active line of research focuses on overcoming the generalization gap (accuracy loss).
Hoffer~\etal{}~\cite{hoffer2017train} suggest to train more epochs, however, training more epochs consumes more time. 
Some heuristic techniques were proposed to close the gap without prolonging epochs. Those techniques include linear learning rate scaling \cite{goyal2017accurate}, warm-up training \cite{goyal2017accurate}\cite{akiba2017extremely}, Layer-wise Adaptive Rate Scaling \cite{you2017scaling} and others \cite{jia2018highly}. 
However, without theoretical support, it is unclear to what extent those methods can generalize. 
For example, linear learning rate scaling and warm-up training cannot generalize to CIFAR-10 \cite{hoffer2017train} and other architectures on ImageNet \cite{you2017scaling}.
Compared with those techniques, our \textit{SmoothOut} is an interpretable solution supported by the ``sharp minima'' hypothesis.
More importantly, our experiments show that \sout{} can further improve the accuracy when combined with those state-of-the-art techniques.

\section{\textit{SmoothOut}: Principles, Theory and Implementation}
\label{sec:method}

\begin{figure*}
	\centering
	\includegraphics[width=1.0\textwidth]{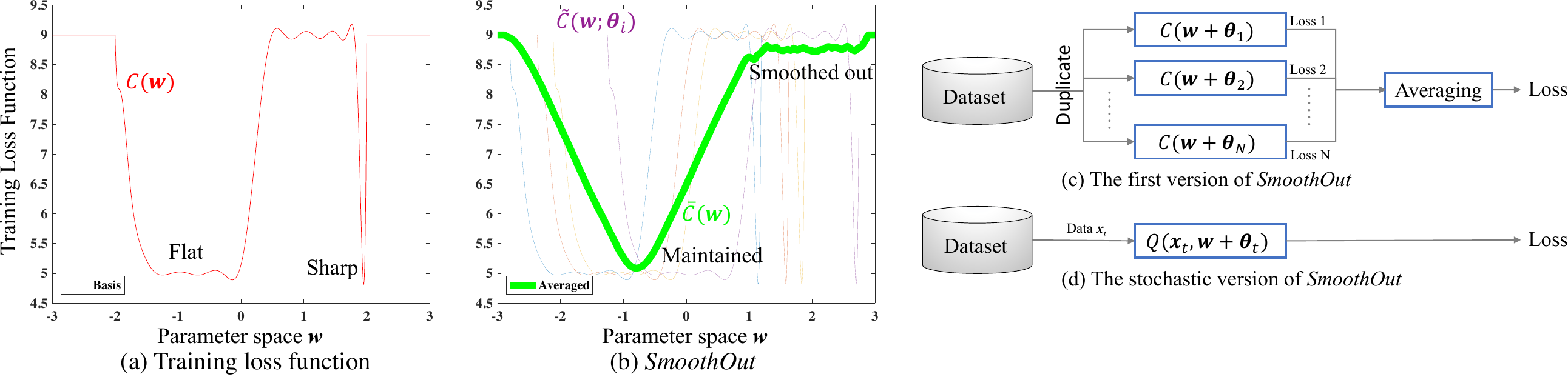}
	\caption{Illustration and framework of \sout{}. (a) The training loss function of the basis model w.r.t. parameter $\bm{w}$. (b) Each thin curve represents a perturbed model; there are totally $1024$ perturbations, but only four are plotted for cleaner visualization; the perturbation is done by slightly shifting the basis model in (a); the shift distance is randomly drawn from a uniform distribution; the green curve is a new model by averaging over all perturbations.
		(c) The first version of proposed \sout{} framework in Eq.~(\ref{eq:first_sout}). (d) The \textit{Stochastic} \sout{} which randomly perturbs parameter $\bm{w}$ at each batch.}
	\label{fig:showcase}
\end{figure*}

We first introduce our \sout{} method and its principles in Section~\ref{sec:method:principle}.
To reduce computation complexity, \textit{Stochastic} \sout{} is proposed in Section~\ref{sec:method:theory}; we prove that \textit{Stochastic} \sout{} is an unbiased approximation of deterministic \sout.
Section~\ref{sec:method:implementation} implements \textit{Stochastic} \sout{} in back-propagation of DNNs. At last, an adaptive variant -- \textit{Ada}\sout{}, is introduced.

\subsection{Principles: Averaging Perturbed Models Smooths Out Sharp Minima}
\label{sec:method:principle}

As \cite{keskar2016large} studied, sharp minima have large generalization gaps, because small distortion/shift of testing function from training function can significantly increase testing loss even though current parameter is a minimum of the training function\footnote{Figure 1 in their paper \cite{keskar2016large} illustrates this conception.}.
Our optimization goal is to encourage convergence to flat minima for more robust models.
Our solution is derived from the sensitivity nature of sharp minima.
We intentionally inject noises into the \textit{model} to smooth out sharp minima. 
The concept is illustrated in Figure~\ref{fig:showcase}(a)(b). 
We define $\bm{w}$ as a point in the parameter space, $C(\bm{w})$ as the training loss function and $\tilde{C}(\bm{w};\bm{\Theta})$ as a perturbation of $C(\bm{w})$. $\tilde{C}(\bm{w};\bm{\Theta})$ is parameterized by both $\bm{w}$ and $\bm{\Theta}$, where $\bm{\Theta}$ is a random vector to generate the perturbation.
Instead of minimizing $C(\bm{w})$, we propose to minimize 
\begin{equation}
    \label{eq:first_sout}
	\bar{C}(\bm{w}) = \textbf{E} \left\lbrace \tilde{C}(\bm{w};\bm{\Theta}) \right\rbrace 
	\approx \frac{1}{N} \sum_{i=1}^{N} \tilde{C}(\bm{w}; \bm{\theta}_i)
\end{equation}
to find a optimal $\bm{w}^*$ for $C(\bm{w})$, where $\bm{\theta}_i$ is a sample of $\bm{\Theta}$ and $N$ is the number of samples.
For simplicity, we assume $C(\bm{w})$ has one flat minimum $\bm{w}_f$ and one sharp minimum $\bm{w}_s$, but the discussion can be generalized to $C(\bm{w})$ with multiple flat and sharp minima.
Our goal is to design an auxiliary function $\bar{C}(\bm{w})$ such that its minimum within the original flat region can approximate $\bm{w}_f$, by satisfying the \textit{Flat Constraint}
\begin{equation}
\label{eq:cgoal1}
\left| \underset{\bm{w} \in \df{}}{ \arg\min } \left( \bar{C}(\bm{w}) \right) - \bm{w}_f \right| \leq \varphi,
\end{equation}
meanwhile the sharp $\bm{w}_s$ is smoothed out, by satisfying the \textit{Sharp Constraint}
\begin{equation}
\begin{split}
\label{eq:cgoal2}
\underset{\ds{}}{\min} \left(  \bar{C}(\bm{w}) \right) 
& \geq \underset{\ds{}}{\max} \left(  C(\bm{w}) \right) \\
& > \underset{\df{}}{ \min } \left( \bar{C}(\bm{w}) \right),
\end{split}
\end{equation} 
where $\mathcal{D}(\bm{w}, \varsigma)$ represents a region around $\bm{w}$, being constrained as
\begin{equation}
\label{eq:smallbox}
\mathcal{D}(\bm{w}, \varsigma) = \left\lbrace \bm{w}' \in \mathbb{R}^m: | ( \bm{w}' - \bm{w} )_i| \leq \varsigma, \forall i \in \{1...m\}  \right\rbrace\\.
\end{equation}
When $\varphi$ is small and $\tau$ is large, Inequality~(\ref{eq:cgoal1}) ensures that the auxiliary function $\bar{C}(\bm{w})$ maintains the minimality of $C(\bm{w})$ in the flat region; in the extreme case of $\varphi = 0 $ and $\tau \rightarrow \infty$, the minimum of $\bar{C}(\bm{w})$ is exactly $\bm{w}_f$.
Conversely, near the original sharp region, Inequality~(\ref{eq:cgoal2}) ensures that minimality of $\bar{C}(\bm{w})$ is eliminated when $\varepsilon$ is relatively large, because ${\max}_{\ds{}} \left(  C(\bm{w}) \right)$,  the lower bound of $\bar{C}(\bm{w})$, increases rapidly by slightly increasing $\varepsilon$ around the sharp minimum; in the extreme case of $\varepsilon \rightarrow \infty$, the lower bound is the maximum of $C(\bm{w})$.
In a nutshell, a good design of $\bar{C}(\bm{w})$ allow a small $\varphi$, a large $\tau$  and a large $\varepsilon$.
In this way, minimization process of $\bar{C}(\bm{w})$ will skip $\bm{w}_s$ and converge to $\bm{w}_f$.
It is infeasible to find an optimal $\bar{C}(\bm{w})$ which minimizes $\varphi$ and maximizes $\tau$  and $\varepsilon$, especially when $C(\bm{w})$ is a deep neural network. However, we find that, under the \textit{Uniform Perturbation}
\begin{equation}
\begin{split}
\label{eq:simple_c}
& \tilde{C}(\bm{w};\bm{\Theta}) = C(\bm{w}+\bm{\Theta}) \\ 
& \text{where}~\Theta_i \overset{\text{i.i.d.}}{\sim} U(-a, a)~\text{and}~\forall i \in \{1,2,...,m\},
\end{split}
\end{equation}
$\bar{C}(\bm{w})$ can well perform the purpose. $U(-a, a)$ is a uniform distribution within a range of $[-a,a]$. In this case, we have abused $\bar{C}(\bm{w};a)$ as $\bar{C}(\bm{w})$ in the notation for simplicity.
In the Appendix~\ref{supp:proof_uniform_avg}, we prove that, under \textit{Uniform Perturbation}, appropriate $\varphi$, $\tau$  and $\varepsilon$ can be found to satisfy \textit{Flat Constraint} and \textit{Sharp Constraint}:
\begin{theorem}
	\label{theorem1}
	When $C(\bm{w})$ is symmetric in $\df{}|_{\tau>a}$, the minimum of $\varphi$ is $0$ to satisfy the Flat Constraint when $\bar{C}(\bm{w})$ is generated under the Uniform Perturbation.
\end{theorem}
\begin{theorem}
	\label{theorem2}
	Suppose $C(\bm{w})$ is high dimensional ($\bm{w} \in \mathbb{R}^m, m \rightarrow \infty$) and is symmetric and strictly monotonic in $\mathcal{D}(\bm{w}_s,b)|_{b>a}$, then $\exists a$ such that Sharp Constraint is satisfied with $\varepsilon \rightarrow a^-$ when $\bar{C}(\bm{w})$ is generated under the Uniform Perturbation.
\end{theorem}
In theorems, the symmetry is assumed only near minima, and the loss surface does not have to be symmetric in the whole space. By referring to the visualization of loss landscapes of neural nets in~\cite{li2017visualizing}, it is reasonable to make this assumption near minima.

Besides the rigor proof in the Appendix~\ref{supp:proof_uniform_avg}, \sout{} can be explained from the perspective of signal processing: imagining the parameter space as a time domain and the function as signals, then averaging is a low-pass filter which eliminates high-frequency signals (sharp regions) while maintains low-frequency signals (flat regions).


Figure~\ref{fig:showcase}(c) illustrates the framework of the proposed \sout{} in SGD.
All models share the same parameter $\bm{w}$. Before training starts, the $i$-th model is independently perturbed by $\bm{\theta}_i$; 
during training, all $\bm{\theta}_i$ are fixed and an identical batch of data is sent to all models for training.
Because a large $N$ is required for approximation in Eq.~(\ref{eq:first_sout}), the computation complexity and memory usage will be very high, especially when $C(\bm{w})$ is a deep neural network. In the next section, the \ssout{} will be proposed to solve this issue.

\subsection{Theory: \ssout{} is Unbiased}
\label{sec:method:theory}

\begin{figure}
	\centering
	\includegraphics[width=.85\columnwidth]{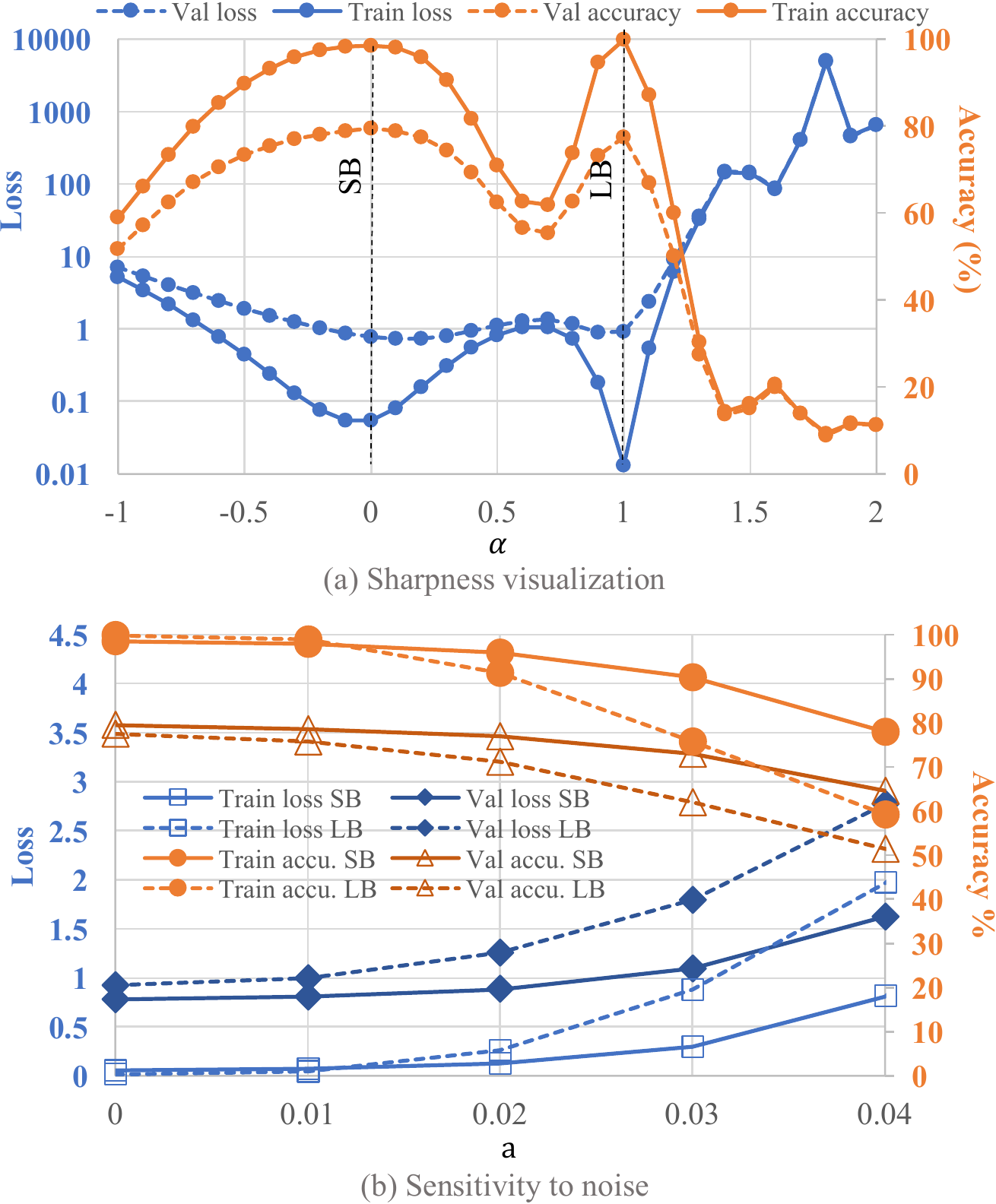}
	\caption{Notation:``SB'': Small Batch ($256$); ``LB'': Large Batch ($5000$); ``accu.'': accuracy.
		(a) loss and accuracy vs. $\alpha$, which controls $\bm{w}$ along the direction from SB minimum ($\bm{w}_f$) to LB minimum ($\bm{w}_s$);
		(b) loss and accuracy under influence of different strengths of noise. Dataset: CIFAR-10. Network: $C_1$ in \protect\cite{keskar2016large} implemented by \protect\cite{hoffer2017train}. Optimizer: Adam with $0.001$ initial learning rate.}
	\label{fig:cifar10_sharpness_test}
\end{figure}

To reduce the computation complexity and memory usage of \sout{} in Figure~\ref{fig:showcase}(c), \ssout{} is proposed in this section as shown in Figure~\ref{fig:showcase}(d).
Instead of using multiple perturbed models to learn from identical data, only one model is trained.
At the $t$-th batch of training data $\bm{x}_t$, the parameter $\bm{w}_t$ is first perturbed to $\bm{w}_t+\bm{\theta}_t$ and then $\bm{x}_t$ is fed into the model to calculate the loss function. 
We can prove that, in both frameworks, the outputs can approximate $\bar{C}(\bm{w})$ without bias.

Formally, in Figure~\ref{fig:showcase}(c), the expectation of the output is
\begin{equation}
\begin{split}
& \textbf{E}_{\bm{\theta}_{1 \dots N}} \left\lbrace \frac{1}{N} \sum_{i=1}^{N} C(\bm{w}+\bm{\theta}_i) \right\rbrace \\
& = \textbf{E}_{\bm{\Theta}} \left\lbrace C(\bm{w}+\bm{\Theta}) \right\rbrace
= \bar{C}(\bm{w}) .
\end{split}
\end{equation}

In  online learning systems \cite{bottou1998online} like Figure~\ref{fig:showcase}(d), the data $\bm{x}_t$ is independently generated from a random distribution and its online loss is obtained by model $Q(\bm{x}_t, \bm{w})$; the final loss function to minimize is the expectation of online loss under data distribution, \ie{},
\begin{equation}
C(\bm{w}) \triangleq \textbf{E}_{\bm{X}} \left\lbrace Q(\bm{x}, \bm{w}) \right\rbrace.
\end{equation}
Therefore, in Figure~\ref{fig:showcase}(d), the expectation of the output is
\begin{equation}
\begin{split}
\textbf{E} \left\lbrace Q(\bm{x}_t, \bm{w}+\bm{\theta}_t) \right\rbrace & =  \textbf{E}_{\bm{\Theta}} \left\lbrace  \textbf{E}_{\bm{X}} \left\lbrace Q(\bm{x}, \bm{w}+\bm{\Theta}) \right\rbrace  \right\rbrace \\
& = \textbf{E}_{\bm{\Theta}} \left\lbrace C(\bm{w}+\bm{\Theta}) \right\rbrace \\
& = \bar{C}(\bm{w}) .
\end{split}
\end{equation}
Consequently, both frameworks in Figure~\ref{fig:showcase} can approximate $\bar{C}(\bm{w}) = \textbf{E} \{ \tilde{C}(\bm{w};\bm{\Theta}) \} $ in Eq.~(\ref{eq:first_sout}), but \ssout{} is much more computation efficient. The only overhead of \ssout{} is noise injection and denoising as will be shown. 
In the following sections, without explicit clarification, \sout{} will refer to the stochastic version in Figure~\ref{fig:showcase}(d).

The reason why \sout{} can eliminate sharp minima is that $C(\bm{w}_s)$ is more sensitive to noise than $C(\bm{w}_f)$, and we expect $\bar{C}(\bm{w}_s)$ increases faster than $\bar{C}(\bm{w}_f)$ as the noise strength $a$ increases from $0$. 
To verify this, we first train a DNN under a small batch size to get a flat minimum $\bm{w}_f$; 
second, $\bm{w}=\bm{w}_f$ is deployed into the framework in Figure~\ref{fig:showcase}(d); 
third, the whole train/validation dataset is fed to the framework in batch size of $100$, and at each batch, the parameter is perturbed to $\bm{w}=\bm{w}_f+\bm{\theta}_t$; 
finally, the losses are averaged over all batches to estimate $\bar{C}(\bm{w}_f)$. 
The same process is done using a large batch size for the same DNN to estimate $\bar{C}(\bm{w}_s)$.
We scan $a$ in a range to test the sensitivity of $\bar{C}(\bm{w}_f)$ and $\bar{C}(\bm{w}_s)$ to perturbation.
Figure~\ref{fig:cifar10_sharpness_test}(a) visualizes the sharpness of $C(\bm{w})$ around $\bm{w}_f$ and $\bm{w}_s$, using the technique adopted in \cite{keskar2016large} which was originally proposed in \cite{goodfellow2014qualitatively}. 
In~Figure~\ref{fig:cifar10_sharpness_test}(a), each point on the loss curve is $( \bm{w}^+, C(\bm{w}^+))$ where $\bm{w}^+ = \alpha \cdot \bm{w}_s + (1-\alpha) \cdot \bm{w}_f$. The visualization is consistent with \cite{keskar2016large}, which concluded that large-batch training converges to sharp minima. 
Figure~\ref{fig:cifar10_sharpness_test}(b) analyzes the sensitivity.
For both training and validation datasets, $\bar{C}(\bm{w}_s)$ indeed increases faster than $\bar{C}(\bm{w}_f)$ as $a$ increases. The accuracy curves have a similar trend. Sensitivity analyses of more DNNs and more datasets are included in the Appendix~\ref{supp:sen_vi}.
Therefore, a side outcome of this work is that we can use 
\begin{equation}
s = \frac{\Delta \left( \bar{C}(\bm{w}^*; a)  \right)}{\Delta a}
\end{equation}
as a metric to measure the sharpness of $C(\bm{w})$ at minimum $\bm{w}^*$. A larger $s$ means a sharper minimum.

At last, under our framework, we can view Dropout as an noise under Bernoulli distribution adapting its noise strength to the corresponding weight. Concretely, in Figure~\ref{fig:showcase}(d), $\theta_{ti}=-w_i$ with probability $p$ and $\theta_{ti}=0$ with probability $1-p$, where $p$ is the dropout ratio. Under this view, Dropout can fit into our framework, but it cannot guide the convergence to sharp minima, because the strength of noise $\theta_{ti}=-w_i$ is too large.
\subsection{Implementation: Back-propagation with Perturbation and Denoising}
\label{sec:method:implementation}

\begin{algorithm}
	\small
	\SetAlgoNoLine
	\SetKwInOut{Input}{Input}
	\SetKwInOut{Output}{Output}
	
	\Input{Training dataset $\bm{X}$, total iterations $T$, model $Q(\bm{x}, \bm{w})$ with initial parameter $\bm{w}=\bm{w}_0$}
	\Indp
	\begin{algorithmic}[1]
		\FOR{$t \in \{0,\dots,T-1\}$}
		\STATE Randomly sample a batch data $\bm{x}_t$ from $\bm{X}$
		\STATE Perturbation: $\bm{w}_t = \bm{w}_t + \bm{\theta}_t$
		where $\theta_{ti} \overset{\text{i.i.d.}}{\sim} U(-a, a)$
		\STATE Back-propagation: $\bm{g}_t = \frac{\partial Q(\bm{x}_t, \bm{w}_t)}{\partial \bm{w}_t} $
		\STATE Denoising: $\bm{w}_t = \bm{w}_t - \bm{\theta}_t$
		\STATE Updating: $\bm{w}_{t+1} = \bm{w}_t - \eta_t \cdot \bm{g}_t$
		
		\ENDFOR
	\end{algorithmic}
	
	\Indm\Output{}
	\Indp
	The model $Q(\bm{x}, \bm{w})$ with final parameter $\bm{w}=\bm{w}_T$
	\caption{\sout{} in Back Propagation} 
	\label{algo:smoothbp}
\end{algorithm}

\begin{figure}
	\centering
	\includegraphics[width=.75\columnwidth]{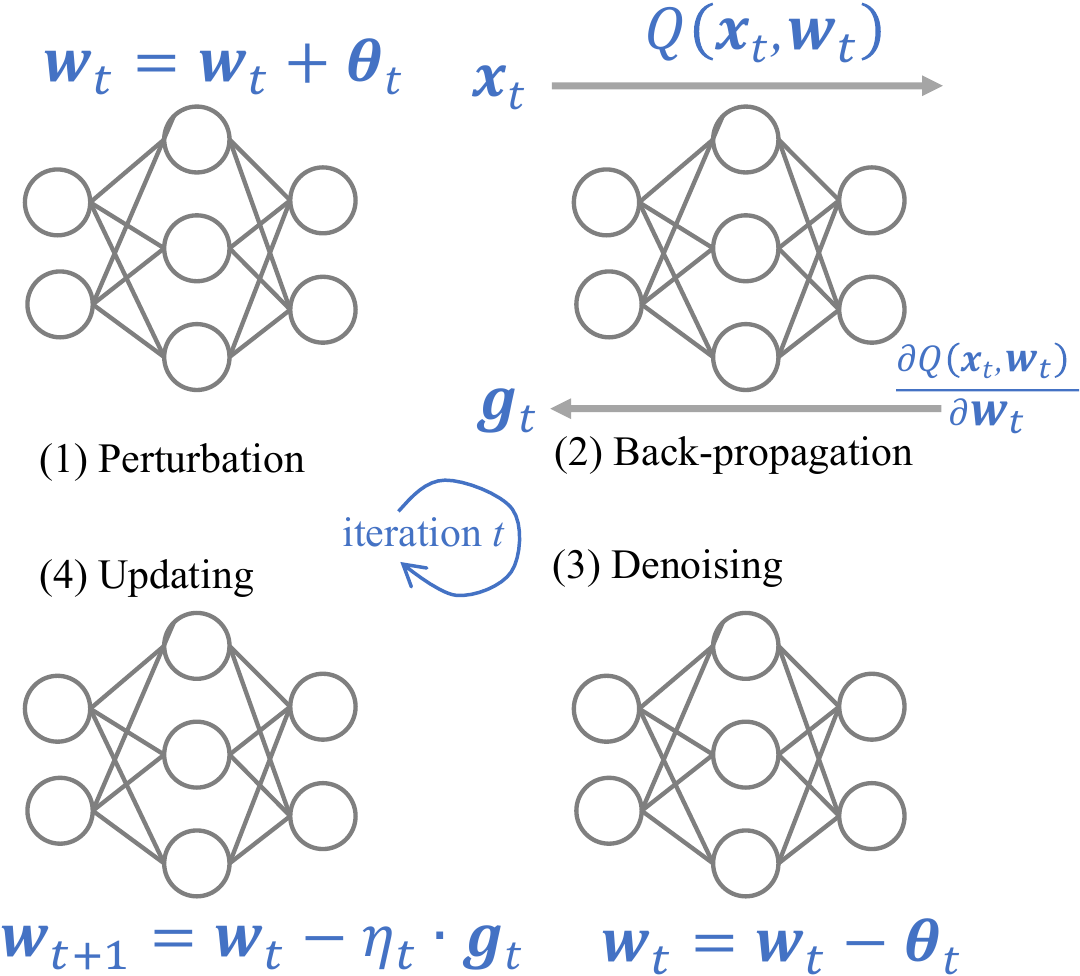}
	\captionof{figure}{\sout{} in BP.}
	\label{fig:smoothbp}
	\vspace{9pt}
\end{figure}

In Figure~\ref{fig:showcase}(d), the gradient to update parameter at iteration $t$ is
\begin{equation}
\begin{split}
\bm{\bm{g}_t} & = \left. \frac{\partial Q(\bm{x}_t, \bm{w} + \bm{\theta}_t)}{\partial \bm{w}} \right|_{\bm{w}=\bm{w}_t} \\
& = \left. \triangledown_{\bm{w}}  Q(\bm{x}_t, \bm{w}) \right|_{\bm{w}=\bm{w}_t + \bm{\theta}_t} \cdot \frac{\partial (\bm{w} + \bm{\theta}_t) }{\partial \bm{w}}.
\end{split}
\end{equation}
Therefore, the parameter is updated as 
\begin{equation}
\bm{w}_{t+1} = \bm{w}_{t} - \eta_t \cdot \left. \triangledown_{\bm{w}}  Q(\bm{x}_t, \bm{w}) \right|_{\bm{w}=\bm{w}_t + \bm{\theta}_t},
\end{equation}
where $\eta_t$ is the learning rate and the gradient is obtained by back propagation when the parameter value is $\bm{w}_t + \bm{\theta}_t$.
Thus, \sout{} can be implemented as Algorithm~\ref{algo:smoothbp} as illustrated in Figure~\ref{fig:smoothbp}. This reveals a pitfall in implementation that the noise $\bm{\theta}_t$ added to $\bm{w}_t$ must be denoised before applying the gradient, which is also a key difference from existing noise injection approaches~\cite{neelakantan2015adding}\cite{mobahi2016training}\cite{fortunato2017noisy}\cite{plappert2017parameter}.

%
%
%

As shown in Figure~\ref{fig:smoothbp}, the only overhead of \sout{} is adding and subtracting noises, which is much more efficient than training multiple DNNs in Figure~\ref{fig:showcase}(c).
Note that, although Algorithm~\ref{algo:smoothbp} is proposed in the context of vanilla SGD, it can be extended to SGD variants by simply utilizing the gradient $\bm{g}_t$ for momentum accumulation, learning rate adaptation, and so on.
\subsection{Adaptive \sout{} -- \textit{AdaSmoothOut}}
\label{sec:method:variants}
%

Due to the fact that the weight distributions across all layers vary a lot, adding noise with a constant strength to all weights may over-perturb the layers with small weights while under-perturb others.
The varying distribution is also the source of problem in visualizing the sharpness as pointed out in \cite{li2017visualizing}. To overcome this, \cite{li2017visualizing} proposed ``filter normalization'' and achieved more accurate visualization.
Inspired by ``filter normalization'', in \sout{}, the noises added to a filter are linearly scaled by $\ell_2$ norm of the filter. In fully-connected layers, the noises are scaled per neuron, \ie{}, all input connections of each neuron form a vector and noises are divided by $\ell_2$ norm of the vector.
We call it \textit{Adaptive} \sout{} (\textit{Ada}\sout{}) because it adapts the strength of noises to the filters instead of fixing the strength. Mathematically, suppose $\bm{w}^{(i)}$ is a vector of parameters in filter $i$ and $\bm{\theta}^{(i)}$ is a noise vector, then adapted noise $\bm{\hat{\theta}}^{(i)}$ will be
\begin{equation}
\bm{\hat{\theta}}^{(i)} = a \cdot \frac{ || \bm{w}^{(i)} ||_2 }{|| \bm{\theta}^{(i)} ||_2} \cdot \bm{\theta}^{(i)},
\end{equation}
where $a$ controls the strength of noises.
Adaptive noise is another key difference from noise injection in previous work. Our ablation study will show adaptive noise is more effective in improving generalization.

\section{Experiments}
\label{sec:exp}

\begin{table*}
  \caption{Sharpness reduction and generalization improvement for DNNs in \protect\cite{keskar2016large}.}
  \label{tab:dnn-smoothout1}
  \centering
  \resizebox{.8\textwidth}{!}{
  	\begin{tabular}{cccccc|r}
  		\toprule
  		DNN & Dataset & Batch size & Baseline  &  \textbf{\sout{}} & Improvement & \begin{tabular}{@{}c@{}}$(\mathcal{C}_{\epsilon}, A)$-sharpness change \\ baseline $\rightarrow$ \textbf{\sout{}} \end{tabular}\\
  		\midrule
  		   \multirow{2}{*}{$F_1$} & \multirow{2}{*}{MNIST} & \multirow{1}{*}{$256$} & $98.49\%$ & $\bm{98.61\%}$ & $0.12\%$ & $0.0000 \rightarrow \bm{0.0000}$ \\
  		   & & \multirow{1}{*}{$6000$} & $98.01\%$ & $\bm{98.42\%}$ & $0.41\%$ & $57.8246 \rightarrow \bm{5.4128}$ \\
  		   \cmidrule{1-7}
  		   
  		   \multirow{2}{*}{$C_1$} & \multirow{2}{*}{CIFAR-10} & \multirow{1}{*}{$256$} & $79.67\%$  & $\bm{81.72\%}$ & $2.05\%$ & $27.7448 \rightarrow \bm{0.0003}$ \\
  		   & & \multirow{1}{*}{$5000$} & $77.30\%$ & $\bm{80.34\%}$ & $3.04\%$ & $117.8266 \rightarrow \bm{0.0004}$ \\
  		   \cmidrule{1-7}
  		   
  		   \multirow{2}{*}{$C_3$} & \multirow{2}{*}{CIFAR-100} & \multirow{1}{*}{$256$} & $47.99\%$ & $\bm{51.17\%}$ & $3.18\%$ & $23.5103 \rightarrow \bm{0.0001}$ \\
  		   & & \multirow{1}{*}{$5000$} & $44.37\%$ & $\bm{48.43\%}$ & $4.06\%$ & $62.0682 \rightarrow \bm{10.7303}$ \\
  		   

  		\bottomrule 	
  		
  	\end{tabular}
  }
\end{table*} 

We evaluate \sout{} in MNIST \cite{lecun1998gradient}, CIFAR-10 \cite{krizhevsky2009learning}, CIFAR-100 \cite{krizhevsky2009learning} and ImageNet \cite{deng2009imagenet} dataset.
\sout{} and \textit{Ada}\sout{} are evaluated in small-batch (``SB'') SGD and large-batch (``LB'') SGD. 
$(\mathcal{C}_{\epsilon}, A)$-sharpness \cite{keskar2016large} is utilized to measure the sharpness of a minimum, which is solved using L-BFGS-B algorithm \cite{byrd1995limited}. In solving $(\mathcal{C}_{\epsilon}, A)$-sharpness, the full-space (\ie{}, $A=I_n$) in the bounding box $\mathcal{C}_{\epsilon}$ (with $\epsilon = 5 \cdot 10^{-4}$) is explored to find the maximum for measurement. As L-BFGS-B is an estimation algorithm and may fail to find the exact maximum value, variance in measurements is observed. We run $5$ experiments for each measurement, and use the \textit{maximum} as the final sharpness metric. Unlike \cite{keskar2016large} which \textit{averaged} over $5$ runs, ours is more reasonable because $(\mathcal{C}_{\epsilon}, A)$-sharpness is based on measuring the \textit{maximum} value around the box. 
In training with \sout{}, 
$a$ is the only additional hyper-parameter to tune, which controls the strength of noise. 
$a$ is very robust because of the width of flat minima. More concretely, $a$ is $0.0375$ in all experiments of \sout{} in Table~\ref{tab:dnn-smoothout1} and Table~\ref{tab:dnn-smoothout2}. We believe the value of $a$ is network architecture dependent (\ie{},
loss function dependent). We cross-validate it in small-batch SGD and directly use it in large-batch SGD
without further tuning, and it generalizes well and improves accuracy in both small-batch SGD and large-batch
SGD.

\begin{figure}
	\centering
	\includegraphics[width=.75\columnwidth]{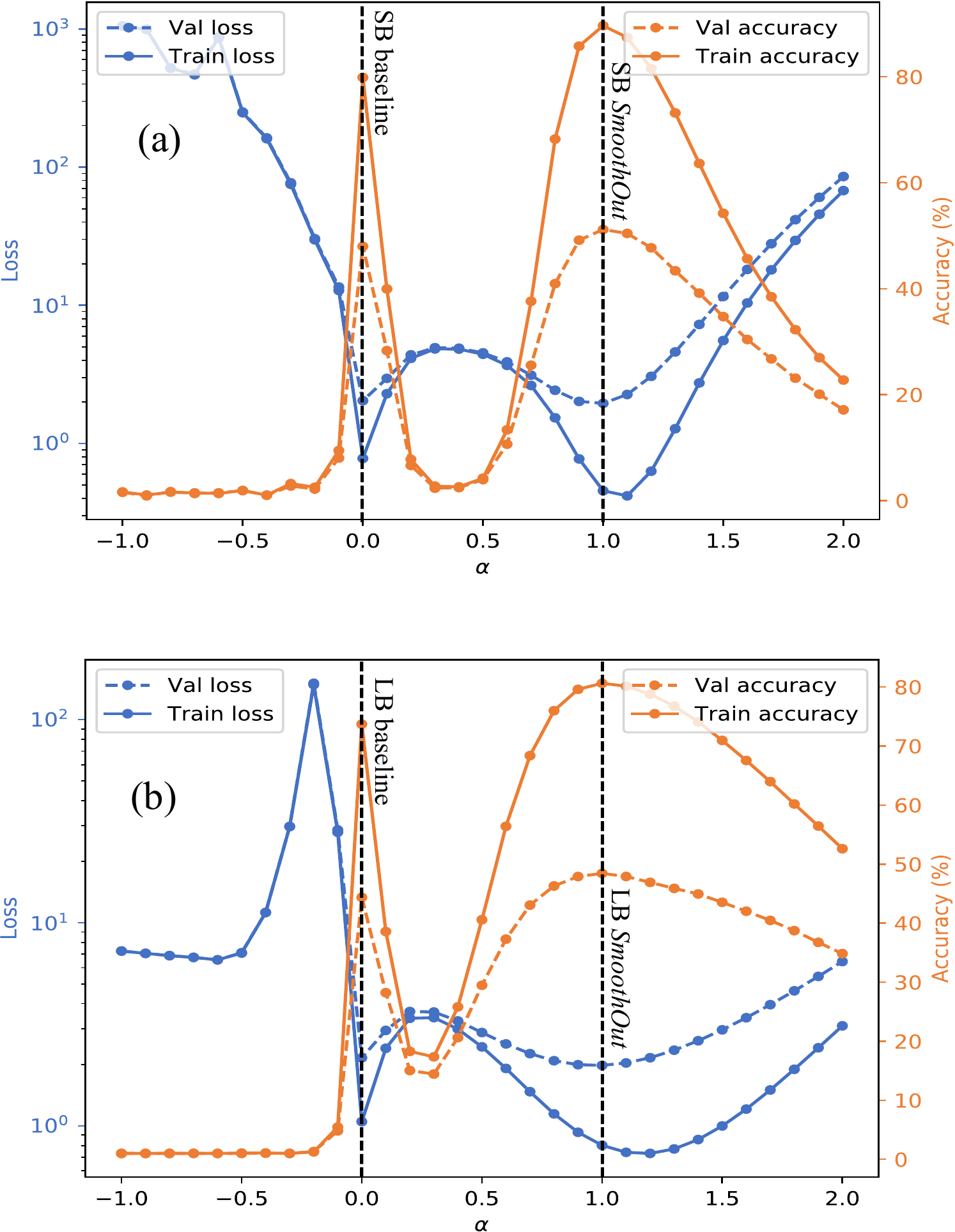}
	\caption{Sharpness of baseline and \sout{} in (a) ``SB'' training and (b) ``LB'' training of $C_3$.}
	\label{fig:cifar100_base_sout}
\end{figure}

\subsection{Convergence to Flatter Minima}
\label{sec:exp-sharpness}


We first adopt benchmarks by \cite{keskar2016large} to verify that \sout{} can effectively guide both SB and LB SGD to flatter minima and thus improve the generalization (accuracy). 
The comparison is in Table~\ref{tab:dnn-smoothout1}. 
Figure~\ref{fig:cifar100_base_sout} visualizes and compares the sharpness of baseline ($C_3$) and \sout{}. Similar visualization results for $F_1$ and $C_1$ can be found in the Appendix~\ref{supp:sen_vi}.
Note that Keskar~\etal{}~\cite{keskar2016large} did not target on achieving state-of-the-art accuracy but studying the characteristics of minima, and we simply follow this purpose. Comparison in state-of-the-art models will be covered in Section~\ref{sec:exp-close-gap}.

In Table~\ref{tab:dnn-smoothout1} and Figure~\ref{fig:cifar100_base_sout}, we observed consistency among sharpness, visualization, and generalization, that is, a smaller $(\mathcal{C}_{\epsilon}, A)$-sharpness, then a flatter region in the visualization and a higher accuracy.
More importantly, the results indicate that 
(1) comparing with SB training, LB training converges to sharper minima with worse generalization, but \sout{} can guide it to converge to flatter minima and closes the gap or even improves the accuracy;
(2) the sharp minima problem also exist in SB training as shown in Figure~\ref{fig:cifar100_base_sout}(a), but \sout{} can reduce the sharpness and improve the accuracy;
(3) sharp minima problem is severer in LB training such that \sout{} can improve more.

At last, we argue that the convergence of our method is stable although noises are injected; that is, different runs converge to similar accuracy under the same strength of injected noises. More specific, for $C_1$ in Table~\ref{tab:dnn-smoothout1}, accuracy standard deviation
is $\pm 0.33\%$, $\pm 0.12\%$, $\pm 0.24\%$ and $\pm 0.31\%$ in small-batch baseline, small-batch
\sout{}, large-batch baseline and large-batch \sout{}, respectively.
\begin{table*}
  \caption{\sout{} and \textit{Ada}\sout{} improve the state-of-the-art baselines.}
  \label{tab:dnn-smoothout2}
  \centering
  \resizebox{.7\textwidth}{!}{
  	\begin{tabular}{ccccccc}
  		\toprule
  		DNN & Dataset & Batch size & Epochs & LRS & Method  &  Accuracy \\
  		\midrule
  		
  		   
  		   \multirow{3}{*}{\textit{ResNet44}} & \multirow{3}{*}{CIFAR-10} & \multirow{3}{*}{$2048$} & \multirow{3}{*}{$400$} & \multirow{3}{*}{square root} & Baseline & $91.02\%$  \\
  		   & & & & & \sout & $91.95\%$ \\
  		   & & & & & \textit{Ada}\sout & $92.63\%$ \\
  		   \cmidrule{1-7}
  		   
  		   \multirow{9}{*}{\textit{ResNet44}} & \multirow{9}{*}{CIFAR-100} & \multirow{9}{*}{$1024$} & \multirow{3}{*}{$200$} & \multirow{3}{*}{square root} & Baseline & $67.23\%$ \\
  		   & & & & & \sout & $68.68\%$ \\
  		   & & & & & \textit{Ada}\sout & $70.09\%$ \\
  		   \cmidrule{4-7}
  		   
  		   & & & \multirow{3}{*}{$400$} & \multirow{3}{*}{square root} & Baseline & $68.62\%$ \\
  		   & & & & & \sout & $70.01\%$ \\
  		   & & & & & \textit{Ada}\sout & $72.39\%$ \\
  		   \cmidrule{4-7}
  		   
  		   & & & \multirow{3}{*}{$400$} & \multirow{3}{*}{linear} & Baseline & $71.21\%$ \\
  		   & & & & & \sout & $71.67\%$ \\
  		   & & & & & \textit{Ada}\sout & $72.85\%$ \\
  		   \cmidrule{1-7}
  		   
  		   \multirow{4}{*}{\textit{AlexNet}} & \multirow{4}{*}{ImageNet} & \multirow{4}{*}{$16384$} & \multirow{2}{*}{$60$} & \multirow{2}{*}{square root} & Baseline & $47.64\%$ \\
  		   & & & & & \textit{Ada}\sout & $52.53\%$ \\
  		   \cmidrule{4-7}
  		   & & & \multirow{2}{*}{$120$} & \multirow{2}{*}{square root} & Baseline & $54.24\%$ \\
  		   & & & & & \textit{Ada}\sout & $55.51\%$ \\
  		   
  		   \cmidrule{1-7}
  		   \multirow{2}{*}{\textit{ResNet18}} & \multirow{2}{*}{ImageNet} & \multirow{2}{*}{$16384$} & \multirow{2}{*}{$90$} & \multirow{2}{*}{square root} & Baseline & $66.75\%$ \\
  		   & & & & & \textit{Ada}\sout & $67.11\%$ \\
  		   

  		\bottomrule 	
  		
  	\end{tabular}
  }
\end{table*} 

\begin{figure}
	\centering
	\includegraphics[width=.75\columnwidth]{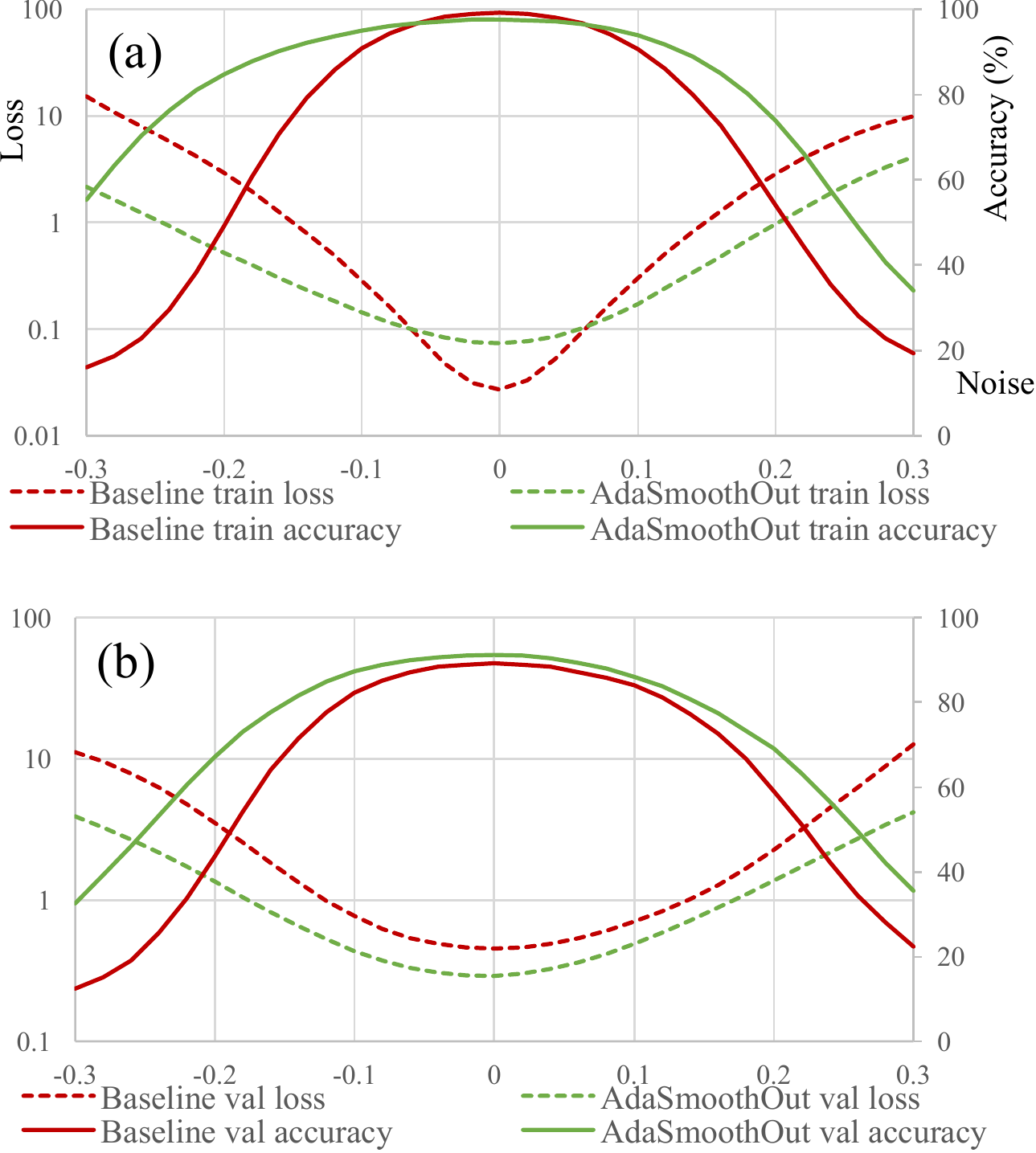}
	\caption{Sharpness visualization by ``filter normalization'' \protect\cite{li2017visualizing} using (a) training dataset and (b) validation dataset. The DNN is \textit{ResNet44} trained by CIFAR-10 using baseline \protect\cite{hoffer2017train} and \textit{Ada}\sout{}.}
	\label{fig:cifar10_norm_visualization}
\end{figure}

\subsection{Improving Generalization on the Top of State-of-the-art Solutions}
\label{sec:exp-close-gap}

In this section, we evaluate our method by state-of-the-art DNNs, including \textit{ResNet44} on CIFAR-10 and CIFAR-100, \textit{AlexNet} \cite{Alex_NIPS2012_4824} and \textit{ResNet18} \cite{Kaiming_ResNet_ICCV} on ImageNet.
Table~\ref{tab:dnn-smoothout2} summarizes all the results.
As generalization issue is severer in LB training, we focus on LB training in this section.
There are lots of proposed techniques to relieve the generalization issue in LB training \cite{hoffer2017train}\cite{goyal2017accurate}\cite{jia2018highly}\cite{akiba2017extremely}\cite{you2017scaling}\cite{smith2017don}, however, our method is orthogonal and we simply apply \sout{} on the top of them to verify if \sout{} can be combined with those state-of-the-art solutions.
We are not able to duplicate all of those techniques, but we select Learning Rate Scaling (LRS), Ghost Batch Normalization (GBN) and Training Longer (TL) techniques \cite{hoffer2017train}\cite{goyal2017accurate} as the representatives.

For LRS, \cite{goyal2017accurate} used linear LRS (\ie{} learning rate is scaled linearly w.r.t. the batch size), while
\cite{hoffer2017train} used square root LRS. The preferable LRS rule is dependent on the dataset and DNN \cite{hoffer2017train}.
In our experiments, linear LRS\footnote{Warm up pre-training is not adopted in our experiments for neat evaluation.} is preferable for \textit{ResNet44} on CIFAR-100 and square root LRS is preferable for the others.
For TL, we simply double the training epochs for each learning rate. 


In the experiments of \textit{ResNet44} on CIFAR-10 and CIFAR-100, we applied GBN, TL ($400$ epochs), linear LRS or square root LRS in the baselines, so that we can diversify the setups to evaluate our method. In all setups, \sout{} improves generalization on the top of the GBN, TL and LRS, verifying that our method is orthogonal to state-of-the-art solutions. 
More importantly, the \textit{Ada}\sout{} variant has the best generalization in all experiments on CIFAR-10 and CIFAR-100, showing the necessity of adaptive noises. Therefore, we choose \textit{Ada}\sout{} as the representative in ImageNet for faster development.

As \textit{Ada}\sout{} is one type of regularizations by stochastic model averaging, the regularizations by weight decay and dropout are not adopted in training ImageNet, such that we can reduce the number of hyperparameters.
The top-1 accuracy of \alexnet{} in SB training is $56.15\%$ with the batch size of $256$, however, in LB training with the batch size of $16384$, the accuracy drops to $47.64\%$ if trained by the same epochs. TL indeed can improve the generalization to $54.24\%$.
More importantly, \textit{Ada}\sout{} improves the accuracy in both cases, \ie{}, improving $4.89\%$ when TL is not applied and improving $1.27\%$ on the top of TL.
Last but not the least, our method also achieve improvement on \textit{ResNet18} on the ImageNet.

At the end, we visualize the sharpness of minima by ``filter normalization'' visualization \cite{li2017visualizing}, \textit{Ada}\sout{} indeed converges to a flatter region as shown in Figure~\ref{fig:cifar10_norm_visualization}.


\subsection{Ablation Study}
\label{sec:exp-ablation}

\subsubsection{The necessity of de-noising}
\begin{table}
  \caption{\sout{} without de-noising tested on CIFAR-10.}
  \label{tab:dnn-no-denoising}
  \centering
  \resizebox{.5\textwidth}{!}{
  	\begin{tabular}{cccc}
  		\toprule
  		DNN & Batch size &  \sout{} &  \sout{} w/o de-noising \\
  		\midrule
  		   
  		   \multirow{2}{*}{$C_1$} & \multirow{1}{*}{$256$} & $81.72\%$  & $36.52\%$ \\
  		   & \multirow{1}{*}{$5000$} & $80.34\%$ & $46.05\%$  \\
  		   $ResNet44$ & $2048$ & $91.95\%$ & $27.57\%$ \\


  		\bottomrule 	
  		\multicolumn{4}{l}{Noise strength $a$ is $0.0375$ in all experiments. }	
  		
  	\end{tabular}
  }
\end{table} 
One of our contributions is the de-noising process. We perform an ablation study by removing the de-noising process to test its necessity.
We rerun all CIFAR-10 \sout{} experiments in Table~\ref{tab:dnn-smoothout1} and Table~\ref{tab:dnn-smoothout2}, but without
de-noising. We use the same noise strength for comparison. The results are summarized in Table~\ref{tab:dnn-no-denoising}.
Without de-noising, the accuracy significantly drops.
The reason is straightforward: strong noises make original parameters and gradients less accurate and deteriorate convergence, 
but our gradients are exactly the gradients of auxiliary function $\bar{C}(\bm{w})$ and perturbed parameters are recovered before applying gradients.

\begin{table}
  \caption{Accuracy with and without de-noising tested by $ResNet44$ on CIFAR-10 with the same setting in Table~\ref{tab:dnn-smoothout2}.}
  \label{tab:dnn-comparison}
  \centering
  \resizebox{.45\textwidth}{!}{
  	\begin{tabular}{ccc}
  		\toprule
  		Method & $a$ & Accuracy \\
  		\midrule
  		Baseline & $0$ & $91.02\%$ \\
  		\midrule
  		   \sout{} & $0.0375$ & $91.95\%$ \\
  		   \sout{} w/o de-noising & $0.0001$ & $91.15\%$ \\
  		   \midrule
  		    \textit{Ada}\sout{} & $0.15$ & $92.63\%$ \\
  		   \textit{Ada}\sout{} w/o de-noising & $0.00075$ & $91.54\%$ \\
  		   

  		\bottomrule 	
  		
  	\end{tabular}
  }
\end{table} 
For a fair comparison, we further carefully tune the noise strength $a$ in \sout{} and \textit{Ada}\sout{} ``w/o de-noising'' to get a near optimal accuracy. More specific, we have to decrease $a$ as SGD is more sensitive to noises when de-noising is not applied.
The results are summarized in Table~\ref{tab:dnn-comparison}. Without de-noising, accuracy is lower.
Note that the de-noising process is naturally generated by our framework and theory in Section~\ref{sec:method}; without de-noising,
it will not fit into our framework and the optimization target will not be the auxiliary function $\bar{C}(\bm{w})$.



\subsubsection{Gaussian noise vs. uniform noise}
Another contribution is the generic \sout{} and \textit{Ada}\sout{} framework, which is agnostic to the type of noises. 
We majorly used uniform noise for study as it is well motivated, but any type of noises can fit into our framework. 
As Gaussian noise is broadly used in the literature~\cite{neelakantan2015adding}\cite{mobahi2016training}\cite{fortunato2017noisy}\cite{plappert2017parameter}\cite{li2016whiteout}\cite{ho2008weight}\cite{kingma2013auto}, we perform an ablation study here by replacing uniform noise with Gaussian noise, for the purpose of verifying our framework is noise agnostic and answering how performance changes when the noise type alters.
The results are in Table~\ref{tab:dnn-uni-vs-gau}, where, in injecting Gaussian noises, $a$ is the standard derivation.
Table~\ref{tab:dnn-uni-vs-gau} indicates that
\begin{itemize}
	\item both uniform and Gaussian noises improve generalization in \sout{} and \textit{Ada}\sout{}, verifying they are agnostic to noise types;
	\item uniform noise is superior to Gaussian noise. 
	A intuitive explanation is that Gaussian distribution gives a high probability to average over values near the minimum, and thus has a smaller probability to smooth out sharp minima. However, uniform distribution evenly treats values around the minimum, and can eliminate the minimum when it is sharp.
	We do not aggressively conclude that uniform noise will always be superior in all settings, but leaving noise selection as an building block when using our framework.
	\item a smaller generalization improvement is observed if only injecting noises into parameters without using our framework (as shown by the ``noise only'' experiments).
	
\end{itemize}

\begin{table}
  \caption{Comparison between uniform and Gaussian noises tested on CIFAR-10 with the same setting in Table~\ref{tab:dnn-smoothout2}.}
  \label{tab:dnn-uni-vs-gau}
  \centering
  \resizebox{.4\textwidth}{!}{
  	\begin{tabular}{ccc}
  		\toprule
  		Method & $a$ & Accuracy \\
  		\midrule
  		Baseline & $0$ & $91.02\%$ \\
  		\midrule
  		   \sout{} (uniform) & $0.0375$ & $91.95\%$ \\
  		   \sout{} (Gaussian) & $0.025$ & $91.53\%$\\
  		\midrule
  		    \textit{Ada}\sout{} (uniform) & $0.15$ & $92.63\%$ \\
  		   \textit{Ada}\sout{} (Gaussian) &  $0.20$ & $92.44\%$ \\
  		\midrule
  		Uniform noise only & $0.0001$ & $91.15\%$ \\
  		Gaussian noise only & $0.00015$  & $91.38\%$\\

  		\bottomrule 	
  		
  	\end{tabular}
  }
\end{table} 

\section{Conclusion}
\label{sec:conclusion}

In this paper, we propose \sout{} and \textit{Ada}\sout{} framework to escape sharp minima during SGD training of Deep Neural Networks, for a better generalization.
\sout{} and \textit{Ada}\sout{} build an auxiliary optimization function without sharp minima, utilizing noise injection. Although noise injection was broadly used in the literature, we interpret the advantage of noise injection from a new perspective of generalization and sharpness. Moreover, our framework advances in multiple ways: (1) de-noising is applied after noise injection; (2) noise strength is adaptive to filter norm in \textit{Ada}\sout{}; (3) uniform noise is majorly adopted for study and can be superior to Gaussian noise in some cases. A comprehensive ablation study is conducted to prove the necessity of those three advances.
In the future, we will extend \sout{} and \textit{Ada}\sout{} to Recurrent Neural Networks, attention-based models and very deep Convolutional Neural Networks.

\appendices

\section{Proof of Theorem \ref{theorem1} and Theorem \ref{theorem2}}
\label{supp:proof_uniform_avg}

Proof of Theorem \ref{theorem1}:

\begin{proof}
	As $\mathcal{D}(\bm{w},a)$ is  defined as a box centering at $\bm{w}$ with size $2a$, \ie{},
	\begin{equation}
	\label{eq:region_a}
	\mathcal{D}(\bm{w},a) = \left\lbrace \bm{w}' \in \mathbb{R}^m: | ( \bm{w}' - \bm{w} )_i| \leq a, \forall i \in \{1...m\} \right\rbrace\\,
	\end{equation}
	then, under \textit{Uniform Perturbation},
	\begin{equation}
	\begin{split}
	\bar{C}(\bm{w}) 
	& = \textbf{E} \left\lbrace \tilde{C}(\bm{w};\bm{\Theta}) \right\rbrace 
	= \textbf{E} \left\lbrace C(\bm{w}+\bm{\Theta}) \right\rbrace \\
	& = \frac{1}{(2a)^m} \underset{\mathcal{D}(\bm{w},a)}{\idotsint} C(\bm{w}') dw'_1 \dots dw'_m 
	\end{split}
	\end{equation}
	
	\begin{equation}
	\begin{split}
	& \frac{\partial \bar{C}(\bm{w}) }{\partial w_i} = \frac{1}{(2a)^m} \cdot \\
	& \underset{\mathcal{D}(\bm{w}_{\textbackslash i},a)}{\idotsint} \left( C(\bm{w}')|_{w'_i=w_i+a} - C(\bm{w}')|_{w'_i=w_i-a} \right) d\bm{w}'_{\textbackslash i},
	\end{split}
	\end{equation}
	where
	\begin{equation}
	\bm{w}_{\textbackslash i} \triangleq [ w_1, \cdots, w_{i-1}, w_{i+1}, \cdots, w_{m}]^T \in \mathbb{R}^{m-1}
	\end{equation} 
	and
	\begin{equation}
	d\bm{w}'_{\textbackslash i} \triangleq dw'_1 \dots dw'_{i-1} dw'_{i+1} \dots dw'_m \cdot
	\end{equation} 
	When $C(\bm{w})$ is symmetric about $\bm{w}_f$ in $\df{}$ such that, $\forall i$, a cut along $w_i = (\bm{w}_f)_i + a$ and a cut along $w_i = (\bm{w}_f)_i - a $ get the same function in the subspace $\bm{w}_{\textbackslash i}$, then $\nabla \bar{C}(\bm{w}_f) = \bm{0}$; that is, the \textit{Flat Constraint} satisfies with $\varphi = 0$. 
\end{proof}
The optimal $\varphi$ and $\tau$ are determined by the symmetry of the flat region. $\varphi$ may be relaxed to a larger value when the symmetry is broken;
however, within a flat region, a larger $\varphi$ may only slightly increase $C(\bm{w}^*)$. 

\vspace{12pt}
Proof of Theorem \ref{theorem2}:

\begin{proof}
	Suppose $C^{(s)}_{\varepsilon'}$ is the maximum value near the sharp minimum, \ie{},
	\begin{equation}
	C^{(s)}_{\varepsilon'} = \underset{\mathcal{D}(\bm{w}_s,\varepsilon')}{\max} \left(  C(\bm{w}) \right),
	\end{equation}
	as $C(\bm{w})$ is strictly monotonic in $\mathcal{D}(\bm{w}_s,b)$, we have, $\forall \varepsilon'<a<b$, 
	\begin{equation}
	\underset{\mathcal{D}(\bm{w}_s,a) \backslash \mathcal{D}(\bm{w}_s,\varepsilon')}{\min} \left(  C(\bm{w}) \right) > C^{(s)}_{\varepsilon'},
	\end{equation}
	where $\mathcal{D}(\bm{w}_s,a) \backslash \mathcal{D}(\bm{w}_s,\varepsilon')$ is a \textit{Set Difference}, notating a domain within $\mathcal{D}(\bm{w}_s,a)$ but outside of $\mathcal{D}(\bm{w}_s,\varepsilon')$.
	
	Then, follow the proof of Theorem~\ref{theorem1}, we have
	\begin{equation}
	\begin{split}
	& \underset{\ds{}|_{\varepsilon<b} }{\min} \left(  \bar{C}(\bm{w}) \right) 
	 = \frac{1}{(2a)^m} \underset{\mathcal{D}(\bm{w}_s,a)}{\idotsint} C(\bm{w}') dw'_1 \dots dw'_m \\
	& \geq  \frac{1}{(2a)^m} \cdot \left( (2a)^m C^{(s)}_{\varepsilon'} - (2\varepsilon')^m \left( C^{(s)}_{\varepsilon'} - C(\bm{w}_s) \right)  \right)\\
	& =  \left( 1 - \left( \frac{\varepsilon'}{a} \right)^m  \right) C^{(s)}_{\varepsilon'} + \left( \frac{\varepsilon'}{a} \right)^m \cdot C(\bm{w}_s).
	\end{split}
	\end{equation}
	Because of
	\begin{equation}
	\lim_{m\to\infty}  \left( \left( 1 - \left( \frac{\varepsilon'}{a} \right)^m  \right) C^{(s)}_{\varepsilon'} + \left( \frac{\varepsilon'}{a} \right)^m \cdot C(\bm{w}_s)  \right)  = C^{(s)}_{\varepsilon'},
	\end{equation}
	in high dimensional models (like deep neural networks), we can find $\varepsilon \rightarrow \varepsilon'^- \rightarrow a^-$ (left limit) to satisfy 
	\begin{equation}
	\label{eq:ieq_bound1}
	\underset{\ds{}}{\min} \left(  \bar{C}(\bm{w}) \right) 
	> C^{(s)}_{\varepsilon} 
	\triangleq \underset{\mathcal{D}(\bm{w}_s,\varepsilon)}{\max} \left(  C(\bm{w}) \right).
	\end{equation}
	In the flat region, 
	\begin{equation}
	\label{eq:ieq_bound2}
	\begin{split}
	\underset{\df{}}{\min} \left(  \bar{C}(\bm{w}) \right) 
	& = \frac{1}{(2a)^m} \underset{\mathcal{D}(\bm{w}_f,a)}{\idotsint} C(\bm{w}') dw'_1 \dots dw'_m \\
	& < \frac{1}{(2a)^m} \cdot \left( (2a)^m \cdot \underset{\mathcal{D}(\bm{w}_f,a)}{\max} \left(  C(\bm{w}) \right) \right)\\
	& = \underset{\mathcal{D}(\bm{w}_f,a)}{\max} \left(  C(\bm{w}) \right) \triangleq C^{(f)}_a
	\end{split}
	\end{equation}
	Assuming $C(\bm{w}_s) \approx C(\bm{w}_f)$, as $a$ grows, $C^{(s)}_{\varepsilon}|_{\varepsilon \rightarrow a^-}$ increases fast in the sharp region while $C^{(f)}_a$ increases slowly in the flat region; therefore, $\exists a$ such that
	\begin{equation}
	\label{eq:ieq_bound4}
	C^{(s)}_{\varepsilon} > C^{(f)}_{a}.
	\end{equation}
	 According to Inequality~(\ref{eq:ieq_bound1})(\ref{eq:ieq_bound2})(\ref{eq:ieq_bound4}),
	\begin{equation}
	\label{eq:ieq_bound3}
	\begin{split}
	\underset{\ds{}}{\min} \left(  \bar{C}(\bm{w}) \right) 
	& > \underset{\mathcal{D}(\bm{w}_s,\varepsilon)}{\max} \left(  C(\bm{w}) \right) \\
	& > \underset{\mathcal{D}(\bm{w}_f,a)}{\max} \left(  C(\bm{w}) \right) \\
	& > \underset{\df{}}{\min} \left(  \bar{C}(\bm{w}) \right) 
	\end{split}
	\end{equation}
	
	which satisfies the \textit{Sharp Constraint}. 
\end{proof}

\section{Sensitivity Analyses and  Sharpness Visualization}
\label{supp:sen_vi}

\begin{figure}
\centering
\includegraphics[width=1.\columnwidth]{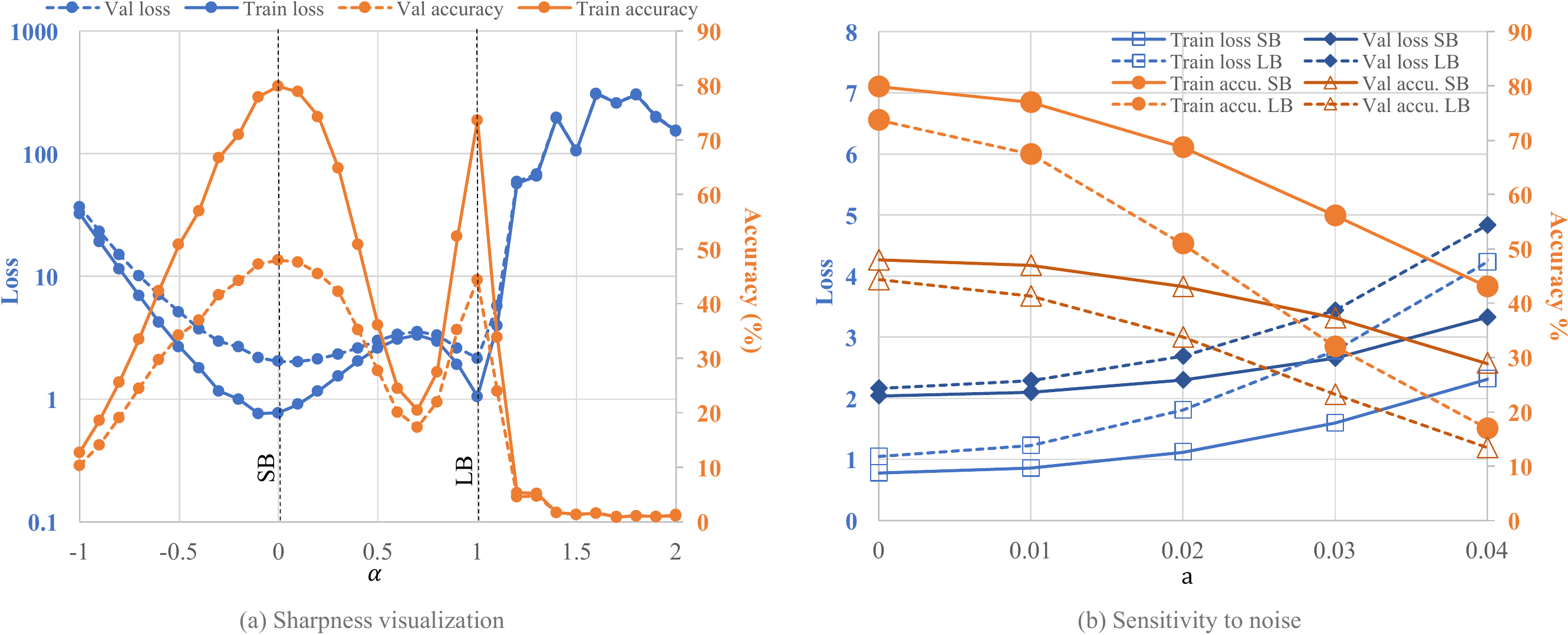}
\caption{Notation:``SB'': Small Batch ($256$); ``LB'': Large Batch ($5000$); ``accu.'': accuracy.
	(a) loss and accuracy vs. $\alpha$, which controls $\bm{w}$ along the direction from SB minimum ($\bm{w}_f$) to LB minimum ($\bm{w}_s$);
	(b) loss and accuracy under influence of different strengths of noise. Dataset: CIFAR-100. Network: $C_3$. The optimizer is Adam with $0.001$ initial learning rate.}
\label{fig:cifar100_sharpness_test}
\end{figure}

\begin{figure}
\centering
\includegraphics[width=1.\columnwidth]{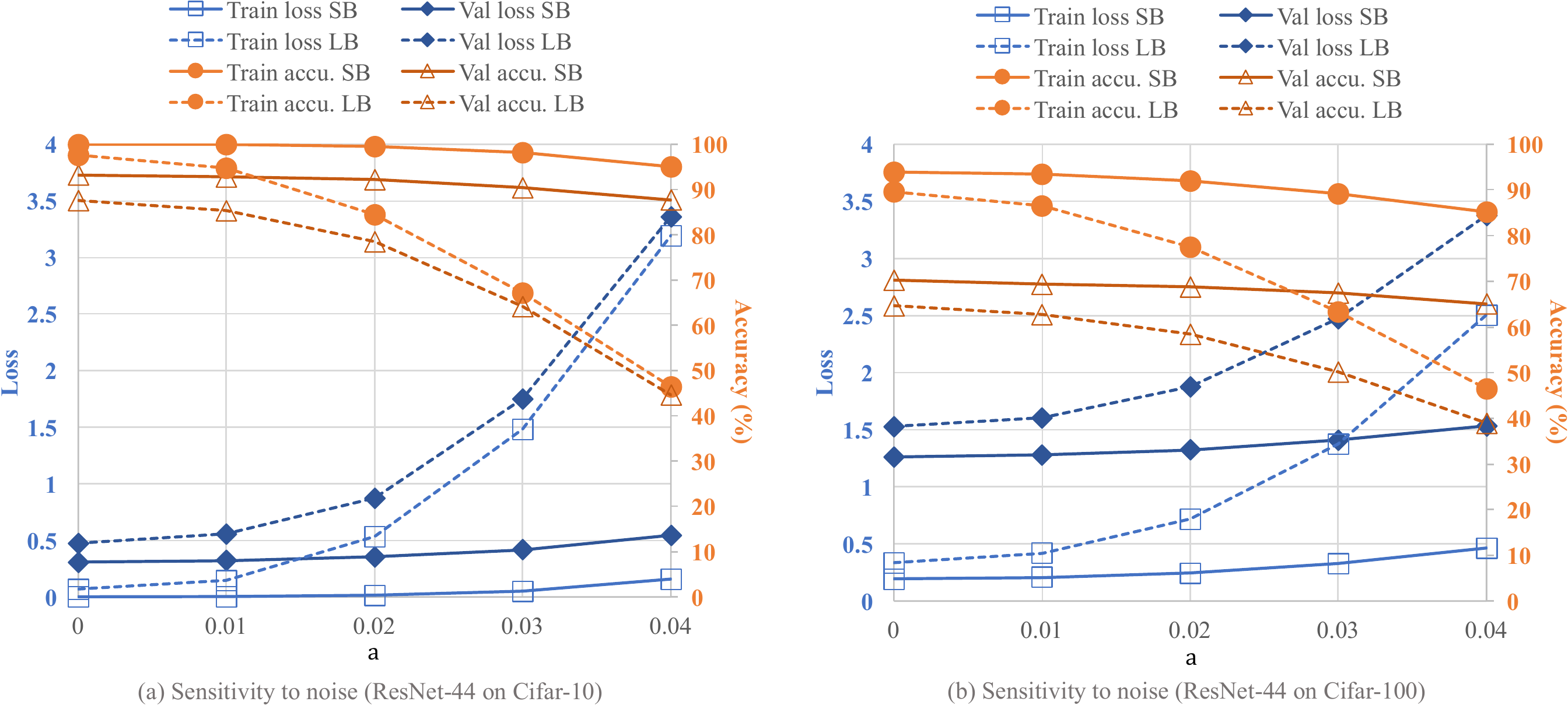}
\caption{
	Loss and accuracy of ResNet-44 under influence of different strengths of noise on (a) CIFAR-10 and (b) CIFAR-100. The optimizer is SGD with momentum $0.9$.
	Notation:``SB'': Small Batch ($128$); ``LB'': Large Batch ($2048$ for CIFAR-10 and $1024$ for CIFAR-100); ``accu.'': accuracy.}
\label{fig:cifar_resnet_sharpness_test}
\end{figure}

\begin{figure}
\centering
\includegraphics[width=1.\columnwidth]{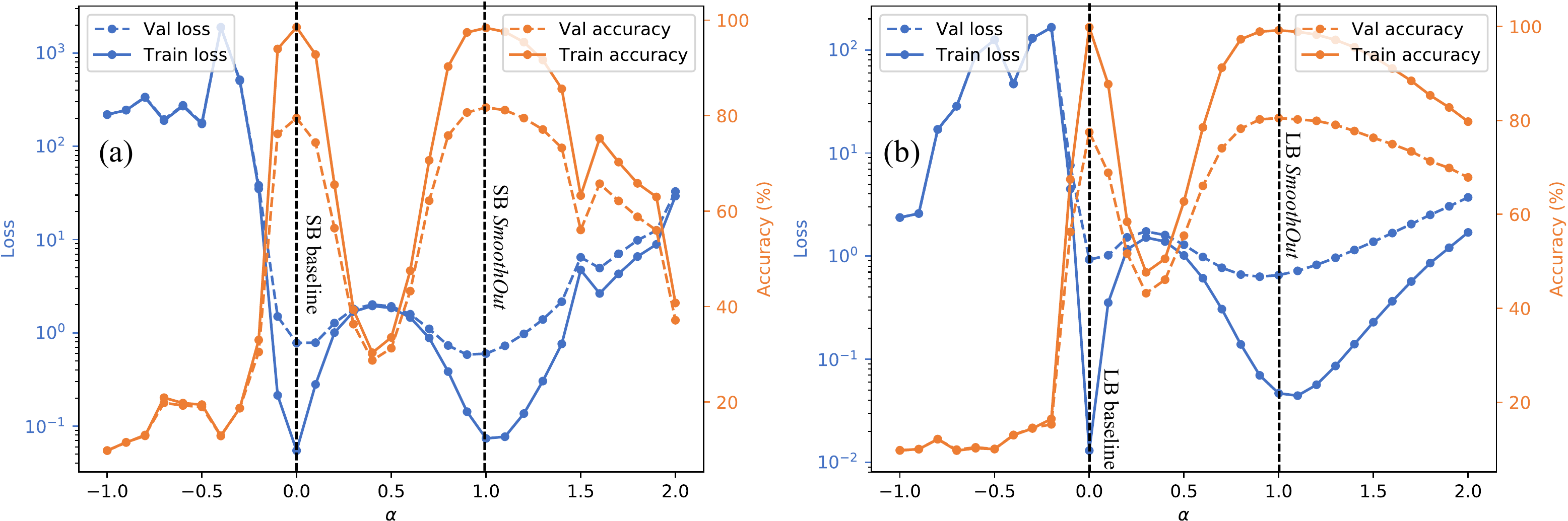}
\caption{Sharpness of baseline and \sout{} in (a) ``SB'' training and (b) ``LB'' training of $C_1$.}
\label{fig:cifar10_base_sout}
\end{figure}

\begin{figure}
\centering
\includegraphics[width=1.\columnwidth]{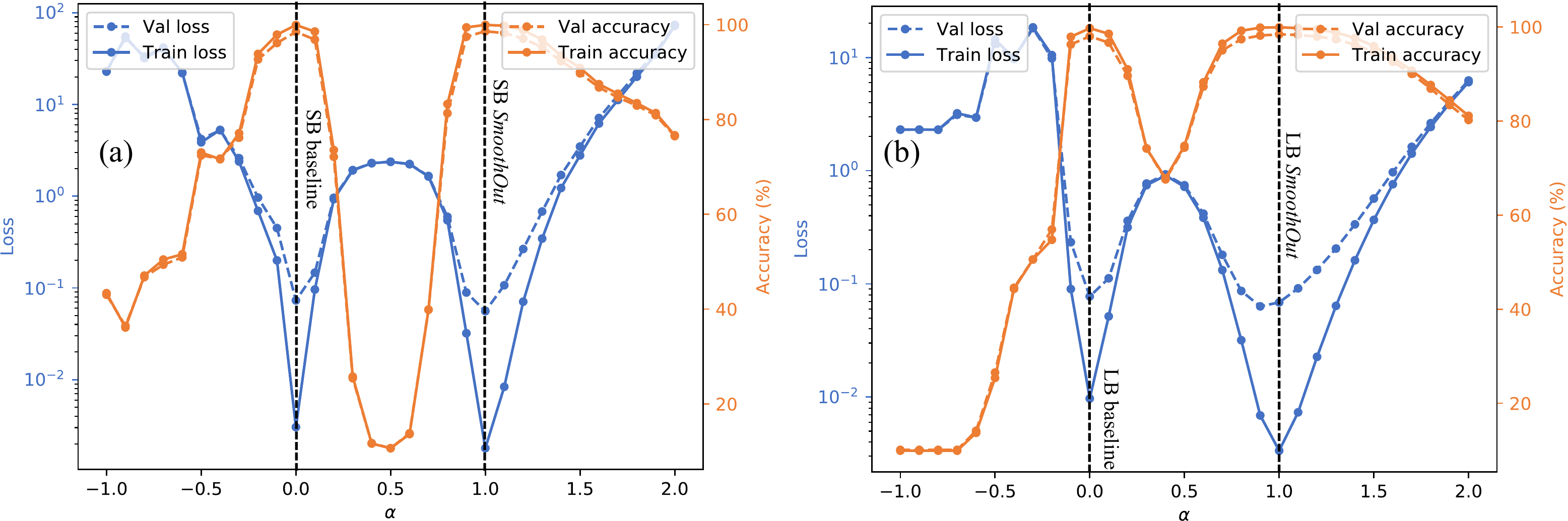}
\caption{Sharpness of baseline and \sout{} in (a) ``SB'' training and (b) ``LB'' training of $F_1$.}
\label{fig:mnist_f1_base_sout}
\end{figure}

We provide more sensitivity analyses in Figure~\ref{fig:cifar100_sharpness_test} and Figure~\ref{fig:cifar_resnet_sharpness_test} as tested on different DNNs and datasets. More sharpness comparison between baseline and \sout{} is visualized in Figure~\ref{fig:cifar10_base_sout} and Figure~\ref{fig:mnist_f1_base_sout}.

\section*{Acknowledgment}
This work was supported in part by DOE DE-SC0018064, NSF 1615475 and 1725456. Any opinions,
findings, conclusions or recommendations expressed in this material are those of the authors and do
not necessarily reflect the views of DOE, NSF or their contractors.

\ifCLASSOPTIONcaptionsoff
  \newpage
\fi



\bibliographystyle{IEEEtran}
\bibliography{tnnls}
\end{document}